\documentclass[10pt,twocolumn,letterpaper]{article}

\usepackage[pagenumbers]{cvpr} 

\usepackage{graphicx}
\usepackage{amsmath}
\usepackage{amssymb}
\usepackage{booktabs}
\usepackage{multirow}
\usepackage[normalem]{ulem}
\useunder{\uline}{\ul}{}
\usepackage{tabularx}
\usepackage[accsupp]{axessibility}  

\usepackage[comma,sort,square,numbers]{natbib}
\usepackage[pagebackref,breaklinks,colorlinks]{hyperref}

\usepackage{bbding}
\usepackage{algorithm}
\usepackage{algorithmic}

\newcommand{\ModelName}{QRNet} 
\newcommand{\AttentionName}{QD-ATT}
\newcommand{\NetworkName}{Query-modulated Refinement Network}
\newcommand{\tabincell}[2]{\begin{tabular}{@{}#1@{}}#2\end{tabular}}

\usepackage{lipsum}

\newcommand\blfootnote[1]{%
  \begingroup
  \renewcommand\thefootnote{}\footnote{#1}%
  \addtocounter{footnote}{-1}%
  \endgroup
}

\usepackage[capitalize]{cleveref}
\crefname{section}{Sec.}{Secs.}
\Crefname{section}{Section}{Sections}
\Crefname{table}{Table}{Tables}
\crefname{table}{Tab.}{Tabs.}

\usepackage{listings}
\usepackage{color}
\usepackage{xcolor}

\definecolor{codegreen}{rgb}{0,0.6,0}
\definecolor{codegray}{rgb}{0.5,0.5,0.5}
\definecolor{codepurple}{rgb}{0.58,0,0.82}
\definecolor{backcolour}{rgb}{1.0,1.0,1.0}

\lstdefinestyle{mystyle}{
    backgroundcolor=\color{backcolour},   
    commentstyle=\color{codegreen},
    keywordstyle=\color{magenta},
    numberstyle=\tiny\color{codegray},
    stringstyle=\color{codepurple},
    basicstyle=\ttfamily\footnotesize,
    breakatwhitespace=false,         
    breaklines=true,                 
    captionpos=b,                    
    keepspaces=true,                 
    numbers=left,                    
    numbersep=5pt,                  
    showspaces=false,                
    showstringspaces=false,
    showtabs=false,                  
    tabsize=2
}

\def\cvprPaperID{719} 
\def\confName{CVPR}
\def\confYear{2022}

\begin{document}

\title{Shifting More Attention to Visual Backbone: Query-modulated Refinement Networks for End-to-End Visual Grounding}

\author{
Jiabo Ye\textsuperscript{1}\quad Junfeng Tian\textsuperscript{2}\quad Ming Yan\textsuperscript{2}\quad Xiaoshan Yang\textsuperscript{3}\\
Xuwu Wang\textsuperscript{4}\quad Ji Zhang\textsuperscript{2}\quad Liang He\textsuperscript{1}\quad Xin Lin\textsuperscript{1}\\
{\small \textsuperscript{1}East China Normal University, Shanghai, China \quad
\textsuperscript{2}Alibaba Group, Hangzhou, China}\\
{\small \textsuperscript{3}NLPR, CASIA, Beijing, China \quad
\textsuperscript{4}Fudan University, Shanghai, China}\\
{\tt\small jiabo.ye@stu.ecnu.edu.cn, \{xlin,lhe\}@cs.ecnu.edu.cn, xwwang18@fudan.edu.cn}\\
{\tt\small xiaoshan.yang@nlpr.ia.ac.cn, \{tjf141457,ym119608,zj122146\}@alibaba-inc.com }
}
\maketitle

\begin{abstract}
   Visual grounding focuses on establishing fine-grained alignment between vision and natural language, which has essential applications in multimodal reasoning systems. Existing methods use pre-trained query-agnostic visual backbones to extract visual feature maps independently without considering the query information. We argue that the visual features extracted from the visual backbones and the features really needed for multimodal reasoning are inconsistent. One reason is that there are differences between pre-training tasks and visual grounding. Moreover, since the backbones are query-agnostic, it is difficult to completely avoid the inconsistency issue by training the visual backbone end-to-end in the visual grounding framework. In this paper, we propose a Query-modulated Refinement Network (QRNet) to address the inconsistent issue by adjusting intermediate features in the visual backbone with a novel Query-aware Dynamic Attention (QD-ATT) mechanism and query-aware multiscale fusion. The QD-ATT can dynamically compute query-dependent visual attention at the spatial and channel levels of the feature maps produced by the visual backbone. We apply the QRNet to an end-to-end visual grounding framework. Extensive experiments show that the proposed method outperforms state-of-the-art methods on five widely used datasets. Our code is available at \url{https://github.com/LukeForeverYoung/QRNet}.

\end{abstract}

\blfootnote{The work was done when Jiabo Ye was working as an intern at Alibaba DAMO Academy.}
\section{Introduction}
\label{sec:intro}
%
%
%
%
Visual grounding~\cite{mao2016generation,yu2016modeling,kazemzadeh2014referitgame,plummer2015flickr30k}, i.e., localizing the referent object in an image according to the given natural language query,
is a fundamental component of multimodal reasoning system.
%
%
Compared with conventional object detection methods~\cite{ren2015faster,redmon2018yolov3} which 
can only recognize the restricted categories contained in the training data,
visual grounding has the advantage of detecting novel combinations of
categories and attributes expressed in free-form text.
In recent years, it has attracted much attention in the field of computer vision and machine learning due to its potential applications in many 
downstream tasks, such as visual question answering~\cite{zhu2016visual7w,gan2017vqs,wang2020general}, visually-grounded language navigation~\cite{anderson2018vision,vogel2010learning} and image captioning~\cite{you2016image,anderson2018bottom,chen2020say}.
\begin{figure}[t]
\centering
\includegraphics[width=1\linewidth]{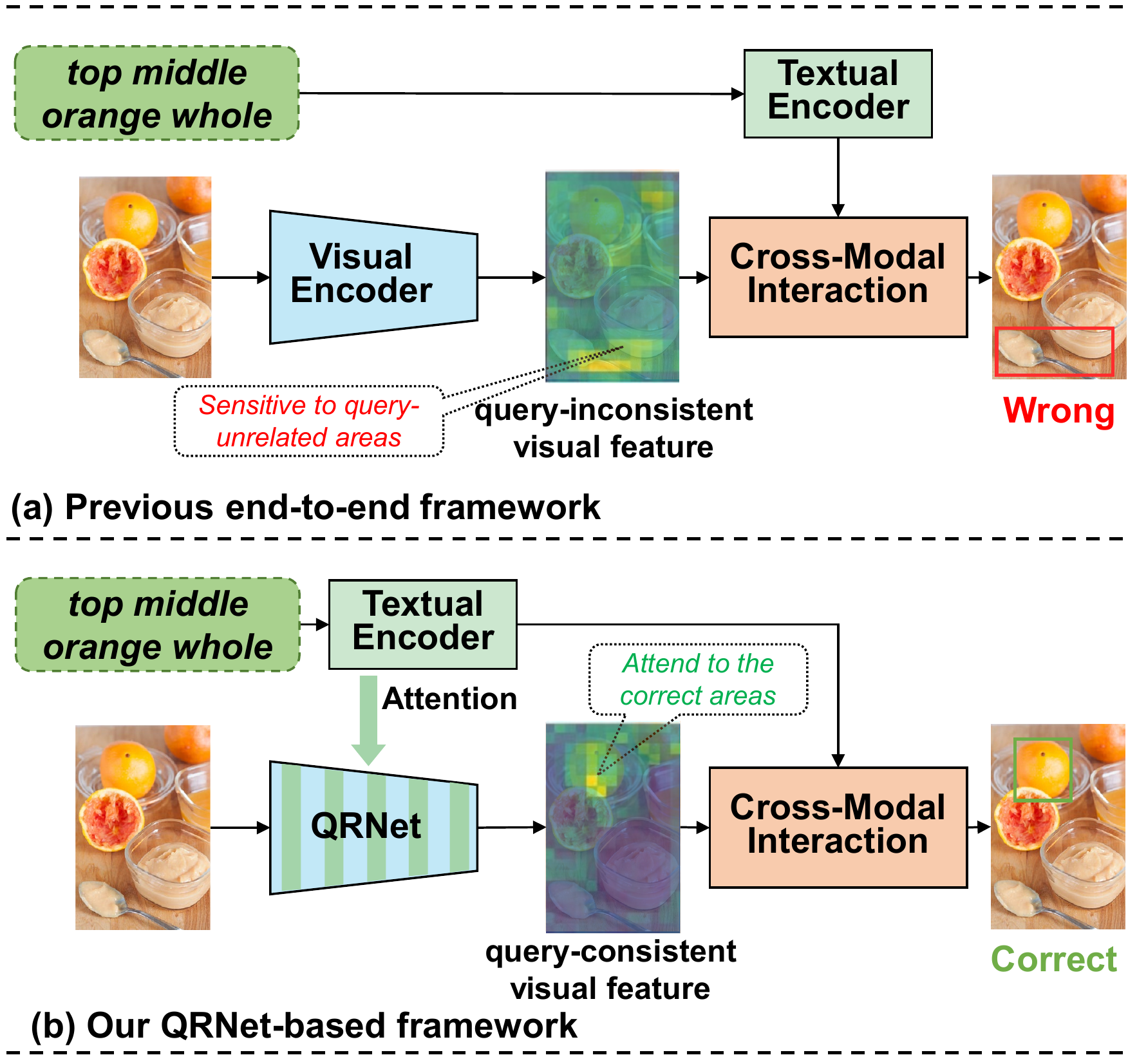}
\caption{(a) A typical end-to-end visual grounding framework that uses two individual encoders to extract visual and textual features for cross-modal interaction. (b) Our visual grounding framework based on a \NetworkName\ (\ModelName).}
\label{fig:title}
\end{figure}

%
The early methods of visual grounding focus on extending
the popularly used one-stage and two-stage object detection architectures.
%
%
One-stage methods~\cite{yang2020improving,ye2021one,deng2021transvg,huang2021look} use a pre-trained fully convolutional network (e.g. Darknet53~\cite{redmon2018yolov3}, ResNet~\cite{he2016deep}) to directly extract pixel-level feature maps and leverage manually-defined dense anchors to return the most likely candidate for the query text.
These methods are easy and efficient for learning or inference, but they cannot perform well on complex queries that have various objects and relations.
Two-stage methods~\cite{yang2019dynamic,yang2020graph,yu2018mattnet} use an off-the-shelf detector (e.g. Faster R-CNN~\cite{ren2015faster}) to extract the region proposals and return the one that best matches the query text using the modality-shared representations.
%
%
These methods always have better performance than the one-stage ones by introducing more complicated multimodal fusion and reasoning mechanisms~\cite{yang2019dynamic,yang2020graph,mu2021disentangled}.
However, the complicated fusion modules
cannot be jointly learned with the detectors, which may
limit their ability in
multimodal
reasoning.
%
More recently,
Transformer~\cite{vaswani2017attention} has been applied in visual grounding~\cite{deng2021transvg,kamath2021mdetr} to conduct the multimodal reasoning more succinctly based on pixel-level feature maps without region proposals or dense anchors.
%
%


%
Although existing visual grounding methods, especially the Transformer-based methods~\cite{deng2021transvg,kamath2021mdetr},
have achieved promising results,
we argue that they do not pay enough attention to the visual backbone which plays a crucial role in effective multimodal reasoning.
Since the visual backbone determines whether all integral visual content in the image is successfully extracted for matching the query text.
Currently, the most widely used backbones are the CNN model (e.g., ResNet~\cite{he2016deep}) pre-trained for image classification on ImageNet and the detector (e.g., Faster R-CNN~\cite{ren2015faster} and Mask R-CNN~\cite{he2017mask}) pre-trained for general object detection of close-set categories.
Therefore, the difference between the visual grounding task and the pre-training task of the backbones may lead to an {\bf inconsistency} between the visual features produced by the backbones and the ones really needed for the multimodal reasoning.
%
%
As shown in \Cref{fig:title}(a), the pre-trained visual backbone extracts general purposed visual features sensitive to regions that may contain the objects of pre-defined categories.
Whereas, the visual grounding requires the backbone to localize a different object referred to by the query.
A straightforward way to alleviate the inconsistency is learning the visual grounding model in an end-to-end form as in~\cite{deng2021transvg}.
%
However,
%
it still cannot completely avoid the inconsistency because the backbone is query-agnostic. In other words, given the same image, the query-agnostic backbone will always output the same feature map no matter what the query sentences are.
%

%
In this paper, we propose a \NetworkName\ (\ModelName) to address the {\bf inconsistency} issue.
%
As shown in \Cref{fig:title}(b), the proposed \ModelName\ can produce query-consistent features by adjusting the feature maps of the visual backbone with the guidance of the query text, which benefits the cross-modal alignment between the query and the relevant region to make a correct prediction.
%
%
%
The \ModelName\ is designed based on Swin-Transformer~\cite{liu2021Swin} and a novel Query-aware Dynamic Attention (\AttentionName),
which can help the \ModelName\ extract query-refined visual feature maps from the visual backbone and fuse the multiscale features with the query guidance.
%
%
The \AttentionName\ dynamically computes textual-dependent visual attentions at the spatial and channel level of the feature maps produced by the visual backbone.
The spatial and channel attentions are further multiplied with the original feature maps to obtain query-refined hierarchical visual feature maps.
%
%
%
%
To comprehensively consider the fine-grained visual features of the candidate regions at different scales,
we aggregate the query-refined visual feature maps obtained at different stages of the \ModelName\ by a query-aware multiscale fusion scheme.
%
%
%
%

%
%
%
%
%

%
%
%
%
%
%

%
We instantiate the proposed \ModelName\ by building a flexible visual grounding framework based on the recently proposed TransVG~\cite{deng2021transvg}. 
We adopt the same multi-layer visual-linguistic  transformer as in~\cite{deng2021transvg} to  perform intra- and  inter-modal reasoning  based on the output token sequence of the \ModelName.
%
 %
The complete pipeline significantly outperforms existing methods, e.g., TransVG~\cite{deng2021transvg} (3.75\% on ReferItGame and 2.85\% on Flickr30K Entities).
%
Note that the proposed \AttentionName\ can be easily applied to other pre-trained visual backbones, e.g.,
ResNet~\cite{he2016deep}.

%

%
%
%

%

%

%

%

The main contributions of this paper are three-fold:
\begin{itemize}
\item
We propose a query-modulated  refinement network to address the inconsistency issue caused by the pre-trained visual backbone through adjusting the visual feature maps with the guidance of query text.
\item 
We propose a novel query-aware dynamic attention mechanism, which can dynamically compute query-dependent spatial and channel attentions for refining visual features.
\item 
We build a flexible visual grounding framework based on the query-modulated  refinement network and demonstrate that it achieves significantly better performance than existing methods on five widely used public datasets.
\end{itemize}

\section{Related Work}
\label{sec:related}
\subsection{Visual Grounding}
The visual grounding methods can be categorized into two-stage methods and one-stage methods.
\textbf{Two-stage methods} divide the visual grounding process into two steps: generation and ranking.
Specifically, a module such as selective search~\cite{uijlings2013selective}, region proposal network~\cite{ren2015faster}, or pre-trained detector~\cite{girshick2014rich,girshick2015fast,ren2015faster} is firstly used to generate proposals that contain objects.
%
%
Then a multimodal ranking network will measure the similarity between the query sentence and proposals and select the best matching result.
Early works~\cite{rohrbach2016grounding,mao2016generation,hu2016natural} only consider sentence-region level similarity.
\citet{yu2018mattnet} decompose the query sentence and image into three modular components related to the subject, location, and relationship to model fine-grained similarity.
Several studies~\cite{bajaj2019g3raphground,yang2019dynamic,yang2020graph,mu2021disentangled} incorporate graph learning to better model cross-modal alignment.
%
%
The more related work to ours is Ref-NMS~\cite{chen2021ref}, which uses the query feature to guide the Non-Maximum Suppression on region proposals to increase the recall of critical objects.
However, Ref-NMS~\cite{chen2021ref} can only be integrated into two-stage methods.
Moreover, it cannot affect the visual backbone in feature extraction and proposal generation.
\textbf{One-stage methods} extract visual feature maps that maintain the spatial structure and perform cross-modal interaction at pixel level.
The fused feature maps are further used to predict bounding boxes.
\citet{yang2019fast} use a Darknet~\cite{redmon2018yolov3} to extract feature maps and broadcast the query embedding to each spatial location.
Several recent works~\cite{liao2020real,yang2020improving,ye2021one} regard multimodal interaction as a multi-step reasoning process to better understand the input query.
\citet{huang2021look} extract landmark features with the guidance of a linguistic description.
The more recent work TransVG~\cite{deng2021transvg} incorporates DETR encoder to extract visual features and proposes a transformer-based visual grounding framework.
\citet{kamath2021mdetr} model the visual grounding as a modulated detection task and propose a novel framework MDETR derived from the DETR detector.
%
%
%
%
%
%
%
%

%

\subsection{Transformer-based Visual Backbones}
\citet{dosovitskiy2021an} propose to extract image features by applying a pure Transformer architecture on image patches.
This method is called Vision Transformer (ViT) and achieves excellent results compared to state-of-the-art convolutional networks.
\citet{pmlr-v139-touvron21a} introduce improved training strategies to train a data-efficient ViT.
\citet{yuan2021tokens} propose a Tokens-to-Token transformation to obtain a global representation and a deep-narrow structure to improve the efficiency.
Some recent works~\cite{chu2021we,han2021transformer} modify the ViT architecture for better performance.
\citet{liu2021Swin} present a novel hierarchical ViT called Swin Transformer.
It computes the representation with shifted windowing scheme to allow for cross-window connection.
Due to the fixed window size, the architecture has linear computational complexity with respect to image size.

\subsection{Multimodal Interaction}
In early studies, multimodal systems use simple interaction methods such as concatenation, element-wise product, summation, and multilayer perception to learn the interaction~\cite{zadeh2017tensor,baltruvsaitis2018multimodal}.
\citet{fukui2016multimodal} introduce a compact bilinear pooling to learn a more complex interaction.
\citet{arevalo2017gated} present a gated unit that can learn to decide how modalities influence the activation.
\cite{nguyen2018improved,nagrani2021attention} use the cross-attention mechanism to enable dense, bi-directional interactions between the visual and textual modalities.
\citet{NIPS2017_6fab6e3a} introduce Conditional Batch Normalization at channel-level to modulate visual feature maps by a linguistic embedding.
\citet{vaezi20mmtm} use squeeze and excitation operations to fuse the multimodal features and recalibrate the channel-wise visual features.
There are also methods~\cite{zadeh2017tensor,baltruvsaitis2018multimodal,arevalo2017gated,vaezi20mmtm,NIPS2017_6fab6e3a} that only perform interaction at channel-level.
However, spatial-level information is also important to some downstream tasks, e.g., visual grounding and visual commonsense reasoning.
Other methods~\cite{fukui2016multimodal,nguyen2018improved,nagrani2021attention} maintain the spatial structure to perform interaction but have a high computational cost.
%
%


\section{Approach}
\label{sec:method}


\subsection{Architecture}
\begin{figure*}[htb]
\includegraphics[width=1\textwidth]{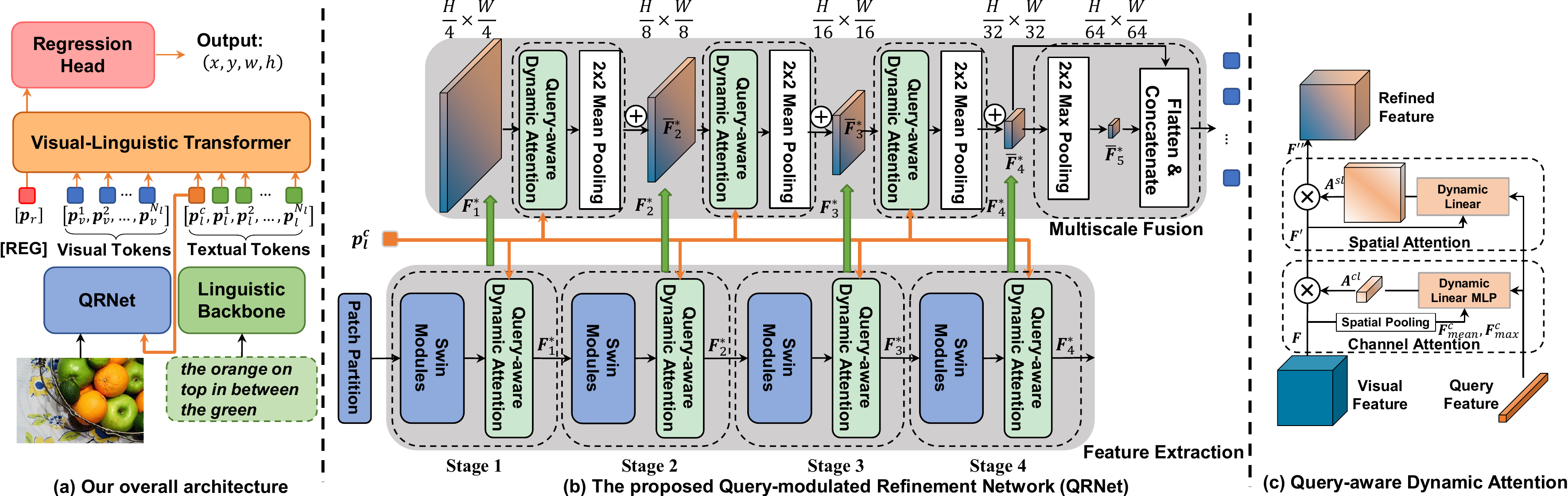}
\caption{(a) The QRNet-based visual grounding framework used in this paper. (b) An overview of our \NetworkName. (c) Illustration of Query-aware Dynamic Attention.}
\label{fig:model}
\end{figure*}
%
In this section, we first formulate the visual grounding task and then present the architecture of the adopted framework.

%
Visual grounding task aims at grounding a query onto a region of an image. The query refers to an object in the image and can be a sentence or a phrase.
It can be formulated as: given an image $I$ and a query $q$, the model needs to predict a bounding box $\mathbf{b}=\{x,y,w,h\}$ which exactly contains the target object expressed by the query.
%

As shown in \Cref{fig:model} (a), 
our framework is designed based on a typical end-to-end visual grounding architecture TransVG~\cite{deng2021transvg}\footnote{https://github.com/djiajunustc/TransVG}.
%
%
%
Given an image and a query sentence, there is a linguistic backbone, typically a pre-trained BERT model, extracting a sequence of 1D feature $\mathbf{T}\in \mathbb{R}^{D_l\times N_l}$.
%
For the visual backbone,
we adopt a new \NetworkName\ (described in \Cref{sec:network})
to extract a flattened sequence of visual feature $\mathbf{V} \in \mathbb{R}^{D_v\times N_v}$.
The major difference between the proposed \NetworkName\ and existing visual backbones is that we take the contextual textual feature from the linguistic backbone to guide the feature extraction and multiscale fusion with the help of Query-aware Dynamic Attention\ (described in \Cref{sec:QDA}).
%
Two projection layers map the visual and linguistic features into the same feature space $\mathbb{R}^{D}$.
The projected visual and linguistic features are denoted by $\mathbf{p}_v\in \mathbb{R}^{D\times N_v}$ and $\mathbf{p}_l\in \mathbb{R}^{D\times N_l}$.
Then $\mathbf{p}_v$ and $\mathbf{p}_l$ are concatenated by inserting a learnable embedding (i.e., a [REG] token) at the beginning of the concatenated sequence.
The joint sequence is formulated as:
\begin{equation}
\begin{aligned}
\boldsymbol{X}=[\mathbf{p}_{r}, \underbrace{\mathbf{p}_{v}^{1}, \mathbf{p}_{v}^{2}, \cdots, \mathbf{p}_{v}^{N_{v}}}_{\text {visual tokens } \boldsymbol{p}_{v}}, \underbrace{\mathbf{p}_{l}^{c}, \mathbf{p}_{l}^{1}, \cdots, \mathbf{p}_{l}^{N_{l}}}_{\text {linguistic tokens } \boldsymbol{p}_{l}}],
\end{aligned}
\end{equation}
where $\mathbf{p}_r$ is the learnable embedding of [REG] token.
$\mathbf{p}_l^c$ is the [CLS] token's representation which is regarded as the contextual textual feature.
Next, a multi-layer visual-linguistic transformer is applied to perform intra- and inter-modal reasoning on the joint sequence.
Finally, a prediction head takes the output representation of the [REG] token to predict the bounding box coordinates $\mathbf{b}$.
The smooth L1 loss~\cite{girshick2015fast} and giou loss~\cite{rezatofighi2019generalized} are used to train the framework.
The training objective can be formulated as:
\begin{equation}
\begin{aligned}
\mathcal{L}=\mathcal{L}_{smooth-l1}{(\mathbf{b},\mathbf{\hat{b}})}+\mathcal{L}_{giou}{(\mathbf{b},\mathbf{\hat{b}})},
\end{aligned}
\end{equation}
where $\mathbf{\hat{b}}$ is the ground-truth box.
%


\subsection{\NetworkName}
\label{sec:network}
%
%
In this section, we introduce the proposed visual backbone of \NetworkName\ (\ModelName).
An overview of the network is shown in \Cref{fig:model} (b).
The network consists of two phases: (1) Query-refined Feature Extraction for extracting query-refined visual feature maps with a hierarchical structure, (2) Query-aware Multiscale Fusion for fusing the extracted feature maps at different scales with the guidance of the query feature.
The two phases both rely on a novel Query-aware Dynamic Attention (\AttentionName), which dynamically computes textual-dependent visual attentions at spatial and channel levels, to enable to compute features with query guidance.
In the following, we will first introduce the implementation of the Query-aware Dynamic Attention.
Then we detail the Query-refined Feature Extraction and Multiscale Fusion.
\subsubsection{Query-aware Dynamic Attention}
\label{sec:QDA}
Now, we will describe the details of our Query-aware Dynamic Attention (shown in \cref{fig:model}(c)).
We first introduce the dynamic linear layer, which learns a linear transformation that can be applied to the visual features to compute the query-aware attentions.
Different from the conventional trainable linear layer,
the parameters of the dynamic linear layer are generated dynamically based on textual features.
%
%
Next, we will illustrate how to use the dynamic linear layer to compute query-aware channel and spatial attentions and obtain the query-consistent visual features. A pseudo-code is presented in the Appendix for better comprehension.
%
%
\paragraph{Dynamic Linear Layer.}
%
Existing visual backbones use modules with static parameters to compute the visual feature maps,
where feeding in the same image will output the same feature maps.
However, in visual grounding, different queries for a single image may reveal different semantic information and intentions, which require different visual features.
We present a dynamic linear layer which can leverage the 
contextual textual feature $\mathbf{p}_l^c\in \mathbb{R}^{D_l}$
%
to guide the mapping from an given input vector $\mathbf{z}_{in}\in \mathbb{R}^{D^{in}}$ to the output $\mathbf{z}_{out}\in \mathbb{R}^{D^{out}}$.
The dynamic linear layer is formalized as follow:
\begin{equation}
\begin{aligned}
\mathbf{z}_{out}&=\mathrm{DyLinear}_{\mathcal{M}_l}(\mathbf{z}_{in})=\mathbf{W}_l^\intercal \mathbf{z}_{in}+\mathbf{b}_l
\end{aligned}
\end{equation}
where $\mathcal{M}_l=\{\mathbf{W}_l,\mathbf{b}_l\},\,\mathbf{W}_l\in \mathbb{R}^{D^{in}\times D^{out}},\, \mathbf{b}_l\in \mathbb{R}^{D^{out}}$.
$D^{in}$ and $D^{out}$ are the dimensions of input and output, respectively.

We use a plain linear layer to generate the $\mathcal{M}_l$.
The generator is denoted by $\mathcal{M}^{'}_l=\Psi(\mathbf{p}_l^c)$, where $\mathcal{M}^{'}_l\in\mathbb{R}^{(D^{in}+1)\ast D^{out}}$.
In detail, we predict a $(D^{in}+1)*D^{out}$ vector which can be reshaped to $\mathcal{M}_l$.
However, it is easy to find that the number of parameters in the generator, i.e., 
$D_l\ast ((D^{in}+1)\ast D^{out})$, is too large.
Such large-scale parameters will slow down the speed of the network and make it easier to be overfitting.
Inspired by matrix factorization, we consider decomposing the $\mathcal{M}_l$ into two factors $\mathbf{U}\in \mathbb{R}^{(D^{in}+1)\times K}$ and $\mathbf{S}\in \mathbb{R}^{K\times D^{out}}$, where $\mathbf{U}$ is a matrix generated from $\mathbf{p}_l^c$, $\mathbf{S}$ is a static learnable matrix, $K$ is a hyper-parameter denoting the factor dimension.
The factor generator $\{\mathbf{W}_l,\mathbf{b}_l\}=\Psi^*(\mathbf{p}_l^c)$ can be formulated as follow:
\begin{equation}
\begin{aligned}
& \mathbf{U}=\mathrm{Reshape}({\mathbf{W}_{g}}^\intercal \mathbf{p}_l^c+\mathbf{b}_{g}),\\
& \mathcal{M}_l=\mathbf{U}\mathbf{S},\\
& \{\mathbf{W}_l,\mathbf{b}_l\}= \mathrm{Split}(\mathcal{M}_l),\\
\end{aligned}
\end{equation}
where $\mathbf{W}_{g}\in \mathbb{R}^{D_l\times (D^{in}+1)\ast K}$ and $ \mathbf{b}_{g}\in \mathbb{R}^{(D^{in}+1)\ast K}$ are trainable parameters of the factor generator.
The parameter matrix $\mathcal{M}_l$ of the dynamic linear layer can be reconstructed by multiplying $\mathbf{U}$ and $\mathbf{S}$.
The $\mathbf{W}_l$ and $\mathbf{b}_l$ can be split from the matrix $\mathcal{M}_l$ along the first dimension.
Finally, we reformulate the dynamic linear layer as follows:
\begin{equation}
\begin{aligned}
\mathbf{z}_{out}&=\mathrm{DyLinear}_{\mathcal{M}_l}(\mathbf{z}_{in})=\mathrm{DyLinear}_{\Psi^*(\mathbf{p}_l^c)}(\mathbf{z}_{in}).
\end{aligned}
\end{equation}

%
Unless otherwise specified, we use $\mathrm{DyLinear}(\mathbf{z}_{in})$ to represent a dynamic linear layer with $\mathbf{p}_l^c$ for simplify. And the different dynamic linear layers do not share parameters.
When the input is a multidimensional tensor, the dynamic linear layer transforms the input at the last dimension.

%
\paragraph{Channel and Spatial Attention.}
%
%
As explained above, the pre-trained visual backbone is sensitive to all the objects it learned in the pre-training task.
%
%
However, only the object which is referred to by the query text is useful.
%
%
Besides, the features useful for the bounding box prediction highly depend on the semantics contained in the query sentence (e.g. entities, descriptions of attributes, and relationships).
In other words, the importance of each channel or each region of the feature map should change dynamically according to the query sentence.
Inspired by Convolutional Block Attention Module~\cite{woo2018cbam}, we consider inferring channel and spatial attentions, i.e., $\mathbf{A}^{cl}$ and $\mathbf{A}^{sl}$, along the separate  dimensions of the feature map to obtain adaptively refined features for a better cross-modal alignment.
%
%

%
Specifically, for a given visual feature map $\mathbf{F}\in \mathbb{R}^{H\times W\times D_v}$, we first aggregate its spatial information by average and maximum pooling and produce $\mathbf{F}^c_{max},\,\mathbf{F}^c_{mean}\in \mathbb{R}^{1\times 1\times D_v}$.
Then, a dynamic multilayer perception consisting of two dynamic linear layers with a ReLU activation in the middle is built to process the pooled features.
The output dimension is the same as the input dimension.
To reduce the number of parameters, we set the dimension of multilayer perception's hidden states as $D_v/r$, where $r=16$ is a reduction ratio.
By feeding the $\mathbf{F}^c_{max}$ and $\mathbf{F}^c_{mean}$ to the dynamic multilayer perception, a Sigmoid function is applied to the summation of the two output features.
The channel attention map $\mathbf{A}^{cl}$ can be captured as follow:
\begin{equation}
\begin{aligned}
\mathbf{F}^{cl}_{mean}&=\mathrm{DyLinear}_{1}(\mathrm{ReLU}(\mathrm{DyLinear}_{2}(\mathbf{F}^c_{mean}))\\
\mathbf{F}^{cl}_{max}&=\mathrm{DyLinear}_{1}(\mathrm{ReLU}(\mathrm{DyLinear}_{2}(\mathbf{F}^c_{max}))\\
\mathbf{A}^{cl}&=\mathrm{Sigmoid}(\mathbf{F}^{cl}_{mean}+\mathbf{F}^{cl}_{max}).
\end{aligned}
\end{equation}
%

%
We perform element-wise multiplication between $\mathbf{F}$ and $\mathbf{A}^{cl}$ to form the channel-wise refined visual feature, where the $\mathbf{A}^{cl}$ is broadcast along the spatial dimension:
\begin{equation}
\begin{aligned}
\mathbf{F}^{'}&=\mathbf{A}^{cl}\otimes \mathbf{F}.
\end{aligned}
\end{equation}

To generate a spatial attention map, instead of squeezing the channel dimension, we leverage another dynamic linear layer to reduce the channel dimension to learn areas of interest to the query and apply an activation function Sigmoid to generate the attention map. 
In short, the computing process is formulated as follows:
\begin{equation}
\begin{aligned}
\mathbf{A}^{sl}&=\mathrm{Sigmoid}(\mathrm{DyLinear}_{3}(\mathbf{F}^{'}))\\
\mathbf{F}^{''}&=\mathbf{A}^{sl}\otimes \mathbf{F}^{'}
\end{aligned}
\end{equation}
where $\mathbf{A}^{sl}\in \mathbb{R}^{H\times W\times 1}$ is the spatial attention map and $\mathbf{F}^{''}$ is the spatially refined visual feature and also the output of our Query-aware Dynamic Attention.

\subsubsection{Query-refined Feature Extraction}
%
%
To extract refined feature maps with the guidance of the query feature, we extend the Swin-Transformer\footnote{https://github.com/SwinTransformer/Swin-Transformer-Object-Detection} to a modulated visual feature extractor.
%
%
%
%
As shown in \Cref{fig:model} (b),
%
%
%
%
%
%
%
%
%
for a given image $I \in \mathbb{R}^{H\times W\times 3}$, 
a patch partition operation is firstly adopted to embed $I$ to $\mathbf{F}_0 \in \mathbb{R}^{\frac{H}{4}\times \frac{W}{4}\times C}$, where $C$ is the embed dimension.
Then $\mathbf{F}_0$ is fed into four cascaded stages,
where each stage consists of multiple Swin-Transformer blocks and a \AttentionName\ module.
In this work, we take the [CLS] representation of the input query sentence from BERT~\cite{devlin2018bert} as the contextual query representation $\mathbf{p}_l^c$ to compute the query-aware dynamic attention.
The $k$-th stage receives the visual feature map $\mathbf{F}_{k-1}^*$ (or $\mathbf{F}_0$ if $k=1$) obtained in the previous stage and generates a transformed feature map $\mathbf{F}_{k}$ through the Swin-Transformer blocks.
%
%
%
Then, the \AttentionName\ module takes the transformed feature $\mathbf{F}_{k}$ and produces a query-aware feature $\mathbf{F}_k^{*}$ that will be further used in the next stage.
%
%
The output of the query-refined feature extractor is a list of hierarchical features $[\mathbf{F}_{1}^{*},\mathbf{F}_{2}^{*},\mathbf{F}_{3}^{*},\mathbf{F}_{4}^{*}]$.
%

\subsubsection{Query-aware Multiscale Fusion}\label{sec:multiscale}

%
Multiscale features are beneficial to detecting objects at different scales~\cite{DBLP:conf/eccv/CaiFFV16}.
However, because visual grounding requires fine-grained interaction after visual backbone (\eg Cross Attention), high-resolution features will dramatically increase the computation.
Therefore, previous works always use the low-resolution features or fuse the multiscale features in a query-agnostic manner,
which will lose scale information or incorporate noise.
Thanks to the hierarchical structure of the modulated Swin-Transformer, we can obtain multiscale features. We fuse the features obtained from different stages with the help of Query-aware Dynamic Attention mechanism and pooling operation.
We flatten and concatenate the low-resolution features as the output token sequence of the backbone.
Specifically, except for the first stage, each stage reduces the resolution of the feature map by a patch merging operator.
In other words, the resolutions of the output feature maps at four stages are $\frac{H}{4}\times\frac{W}{4},\,\frac{H}{8}\times\frac{W}{8},\,\frac{H}{16}\times\frac{W}{16}$ and $\frac{H}{32}\times\frac{W}{32}$, respectively.
Besides, the patch merging operator fuses the patch features into the channel which doubles the channel dimension.
Thus the channel dimensions of the output at the four stages are $C,\,2C,\,4C$ and $8C$, respectively.
Next, we will introduce how to efficiently fuse these multiscale features with the guidance of the query text.
As shown in \Cref{fig:model} (b), we propose to fuse the output features at different stages with the help of \AttentionName.
Specifically, we first use four $1\times 1$ convolutional layers to unify the channel dimensions to $D$.
To filter out noisy signals in the feature maps, we build \AttentionName\ modules for the feature maps $\{\mathbf{F}^{*}_k |\,k=1,2,3 \}$ of the first three stages.
From $\mathbf{F}^{*}_k$, the \AttentionName\ module produces a weighted feature map with the same size as the input.
We apply a $2\times 2$ mean pooling with stride=2 to reduce the resolution to be the same as the next feature map $\mathbf{F}^{*}_{k+1}$ and compute the average of them to obtain $\mathbf{\bar{F}}^{*}_{k+1}$.
Finally, the last feature map $\mathbf{\bar{F}}^{*}_{4}$ contains features of interest to the query from all scales.
To detect very large objects, we also apply a $2\times 2$ max pooling to obtain a $\frac{H}{64}\times\frac{W}{64}$ feature map $\mathbf{\bar{F}}^{*}_{5}$.
We flatten and concatenate the $\mathbf{\bar{F}}^{*}_{4}$ and $\mathbf{\bar{F}}^{*}_{5}$ as the output token sequence $\mathbf{V}$.
%

\section{Experiments}
\label{sec:experiment}

\subsection{Datasets and Evaluation}
We evaluate our method on phrase grounding dataset Flickr30K Entities\cite{plummer2015flickr30k} and referring expression grounding datasets RefCOCO\cite{yu2016modeling}, RefCOCO+\cite{yu2016modeling}, RefCOCOg\cite{mao2016generation} and ReferItGame\cite{kazemzadeh2014referitgame}.
The details and statistics are summarized in Appendix.
In phrase grounding the queries are phrases and in referring expression grounding, they are referring expressions corresponding to the referred objects.
We follow the same metric used in \cite{deng2021transvg}.
Specifically, a prediction is right if the IoU between the grounding-truth box and the predicted bounding box is larger than 0.5.
%

\subsection{Implementation Details}
The QRNet is built based on Swin-Transformer, i.e., Swin-S pre-trained with Mask-RCNN on MSCOCO~\cite{lin2014microsoft}\footnote{We excluded images in the validation and test set of the RefCOCO series from MSCOCO training set and retrained Swin-Transformer for fair comparison.}.
We use $\mathrm{BERT}_{\mathrm{base}}$(uncased)
for linguistic feature extraction.
We set the intermediate dimension $D=256$ and the factor dimension $K=30$.
We follow the TransVG~\cite{deng2021transvg} to process the input images and sentences.
We also follow the training setting used in TransVG, which uses a AdamW optimizer with weight decay $10^{-4}$, sets the batch size to 64 and the dropout ratio to 0.1 for FFN in Transformer.
The learning rate is set to $10^{-5}$ for pre-trained parameters and $10^{-4}$ for other parameters.
The parameters without pre-training are randomly initialized with Xavier.
We train our model for 160 epochs.
The learning rate is multiplied by a factor of 0.1 at epoch 40 for Flickr30K Entities and at epoch 60 for other datasets.
We also follow the commonly used data augmentation strategies in ~\cite{liao2020real,yang2020improving,yang2019fast}.

\subsection{Quantitative Results}
We show the comparison results on ReferItGame and Flickr30k Entities in \Cref{tab:sub_main}.
The ReferItGame collects queries by a cooperative game, requiring one player to write a referring expression for a specified object.
The other one needs to click on the correct object.
Queries in Flickr30k Entities are phrases in the caption.
We observe that the proposed \ModelName\ outperforms previous works.
We replace the visual branch of TransVG with Swin-Transformer and denote it by TransVG(Swin) to explore the impact of backbone.
The performance is similar to the original TransVG, indicating that the accuracy gains are not from Swin-Transformer.
%
\begin{table}[t]
\small
\centering

\begin{tabular}{@{}cccc@{}}
\toprule
\multicolumn{1}{c|}{\multirow{2}{*}{Module}}               & \multicolumn{1}{c|}{\multirow{2}{*}{Backbone}} & \multicolumn{1}{c|}{ReferItGame}    & Flickr30K      \\
\multicolumn{1}{c|}{}                                      & \multicolumn{1}{c|}{}                          & \multicolumn{1}{c|}{test}           & test           \\ \midrule
\multicolumn{4}{l}{\textbf{Two-stage}}                                                                                                                                      \\ \midrule
\multicolumn{1}{c|}{VC~\cite{zhang2018grounding}}           & \multicolumn{1}{c|}{VGG16}                     & \multicolumn{1}{c|}{31.13}          & -              \\
\multicolumn{1}{c|}{MAttNet~\cite{yu2018mattnet}}           & \multicolumn{1}{c|}{ResNet-101}                & \multicolumn{1}{c|}{29.04}          & -              \\
\multicolumn{1}{c|}{Similarity net~\cite{wang2018learning}} & \multicolumn{1}{c|}{ResNet-101}                & \multicolumn{1}{c|}{34.54}          & 50.89          \\
\multicolumn{1}{c|}{LCMCG~\cite{liu2020learning}}           & \multicolumn{1}{c|}{ResNet-101}                & \multicolumn{1}{c|}{-}              & 76.74          \\
\multicolumn{1}{c|}{DIGN~\cite{mu2021disentangled}}         & \multicolumn{1}{c|}{VGG-16}                    & \multicolumn{1}{c|}{{\ul 65.15}}    & {\ul 78.73}    \\ \midrule
\multicolumn{4}{l}{\textbf{One-stage}}                                                                                                                                      \\ \midrule
\multicolumn{1}{c|}{FAOA~\cite{yang2019fast}}               & \multicolumn{1}{c|}{DarkNet-53}                & \multicolumn{1}{c|}{60.67}          & 68.71          \\
\multicolumn{1}{c|}{RCCF~\cite{liao2020real}}               & \multicolumn{1}{c|}{DLA-34}                    & \multicolumn{1}{c|}{63.79}          & -              \\
\multicolumn{1}{c|}{ReSC-Large~\cite{yang2020improving}}    & \multicolumn{1}{c|}{DarkNet-53}                & \multicolumn{1}{c|}{64.60}          & 69.28          \\
\multicolumn{1}{c|}{SAFF~\cite{ye2021one}}                  & \multicolumn{1}{c|}{DarkNet-53}                & \multicolumn{1}{c|}{66.01}          & 70.71          \\
\multicolumn{1}{c|}{TransVG~\cite{deng2021transvg}}         & \multicolumn{1}{c|}{ResNet-101}                & \multicolumn{1}{c|}{70.73}          & {\ul 79.10}    \\
\multicolumn{1}{c|}{TransVG\ (Swin)}                       & \multicolumn{1}{c|}{Swin-S}                    & \multicolumn{1}{c|}{{\ul 70.86}}    & 78.18          \\ \midrule
\multicolumn{1}{c|}{\ModelName\ (ours)}                    & \multicolumn{1}{c|}{Swin-S}                    & \multicolumn{1}{c|}{\textbf{74.61}} & \textbf{81.95} \\ \bottomrule
\end{tabular}
\caption{The performance comparisons (Acc@0.5) on ReferItGame and Flickr30K Entities.}
\label{tab:sub_main}
\end{table}

\begin{table*}[ht]
\centering
\begin{tabular}{@{}ccccccccccc@{}}
\toprule
\multicolumn{1}{c|}{\multirow{2}{*}{Models}}             & \multicolumn{1}{c|}{\multirow{2}{*}{Backbone}} & \multicolumn{3}{c|}{RefCOCO}                                          & \multicolumn{3}{c|}{RefCOCO+}                                         & \multicolumn{3}{c}{RefCOCOg}                     \\
\multicolumn{1}{c|}{}                                    & \multicolumn{1}{c|}{}                          & val            & testA          & \multicolumn{1}{c|}{testB}          & val            & testA          & \multicolumn{1}{c|}{testB}          & val-g          & val-u          & test-u         \\ \midrule
\multicolumn{11}{l}{\textbf{Two-stage}}                                                                                                                                                                                                                                                                               \\ \midrule
\multicolumn{1}{c|}{VC~\cite{zhang2018grounding}}         & \multicolumn{1}{c|}{VGG16}                     & -              & 73.33          & \multicolumn{1}{c|}{67.44}          & -              & 58.40          & \multicolumn{1}{c|}{53.18}          & 62.30          & -              & -              \\
\multicolumn{1}{c|}{MAttNet~\cite{yu2018mattnet}}         & \multicolumn{1}{c|}{ResNet-101}                & 76.65          & 81.14          & \multicolumn{1}{c|}{69.99}          & 65.33          & 71.62          & \multicolumn{1}{c|}{56.02}          & -              & 66.58          & 67.27          \\
\multicolumn{1}{c|}{Ref-NMS~\cite{chen2021ref}}           & \multicolumn{1}{c|}{ResNet-101}                & {\ul 78.82}    & 82.71          & \multicolumn{1}{c|}{{\ul 73.94}}    & 66.95          & 71.29          & \multicolumn{1}{c|}{{\ul 58.40}}    & -              & {\ul 68.89}    & {\ul 68.67}    \\
\multicolumn{1}{c|}{LGRANs~\cite{wang2019neighbourhood}}  & \multicolumn{1}{c|}{VGG16}                     & -              & 76.60          & \multicolumn{1}{c|}{66.40}          & -              & 64.00          & \multicolumn{1}{c|}{53.40}          & 61.78          & -              & -              \\
\multicolumn{1}{c|}{RvG-Tree~\cite{hong2019learning}}     & \multicolumn{1}{c|}{ResNet-101}                & 75.06          & 78.61          & \multicolumn{1}{c|}{69.85}          & 63.51          & 67.45          & \multicolumn{1}{c|}{56.66}          & -              & 66.95          & 66.51          \\
\multicolumn{1}{c|}{CM-Att-Erase~\cite{liu2019improving}} & \multicolumn{1}{c|}{ResNet-101}                & 78.35          & {\ul 83.14}    & \multicolumn{1}{c|}{71.32}          & {\ul 68.09}    & {\ul 73.65}    & \multicolumn{1}{c|}{58.03}          & {\ul 68.67}    & -              & -              \\
\multicolumn{1}{c|}{NMTree~\cite{liu2019learning}}        & \multicolumn{1}{c|}{ResNet-101}                & 76.41          & 81.21          & \multicolumn{1}{c|}{70.09}          & 66.46          & 72.02          & \multicolumn{1}{c|}{57.52}          & 64.62          & 65.87          & 66.44          \\ \midrule
\multicolumn{11}{l}{\textbf{One-stage}}                                                                                                                                                                                                                                                                               \\ \midrule
\multicolumn{1}{c|}{FAOA~\cite{yang2019fast}}             & \multicolumn{1}{c|}{DarkNet-53}                & 72.54          & 74.35          & \multicolumn{1}{c|}{68.50}          & 56.81          & 60.23          & \multicolumn{1}{c|}{49.60}          & 56.12          & 61.33          & 60.36          \\
\multicolumn{1}{c|}{RCCF~\cite{liao2020real}}             & \multicolumn{1}{c|}{DLA-34}                    & -              & 81.06          & \multicolumn{1}{c|}{71.85}          & -              & 70.35          & \multicolumn{1}{c|}{56.32}          & -              & -              & 65.73          \\
\multicolumn{1}{c|}{ReSC-Large~\cite{yang2020improving}}  & \multicolumn{1}{c|}{DarkNet-53}                & 77.63          & 80.45          & \multicolumn{1}{c|}{72.30}          & 63.59          & 68.36          & \multicolumn{1}{c|}{56.81}          & 63.12          & 67.30          & 67.20          \\
\multicolumn{1}{c|}{SAFF~\cite{ye2021one}}                & \multicolumn{1}{c|}{DarkNet-53}                & 79.26          & 81.09          & \multicolumn{1}{c|}{76.55}          & 64.43          & 68.46          & \multicolumn{1}{c|}{58.43}          & -              & 68.94          & 68.91          \\
\multicolumn{1}{c|}{HFRN~\cite{qiu2020language}}          & \multicolumn{1}{c|}{ResNet-101}                & 79.76          & {83.12}    & \multicolumn{1}{c|}{75.51}          & 66.80          & 72.53          & \multicolumn{1}{c|}{59.09}          & -              & {\ul 69.71}    & 69.08          \\
\multicolumn{1}{c|}{ISRL~\cite{sun2021iterative}}         & \multicolumn{1}{c|}{ResNet-101}                & -              & 74.27          & \multicolumn{1}{c|}{68.10}          & -              & 71.05          & \multicolumn{1}{c|}{58.25}          & -              & -              & {\ul 70.05}    \\
\multicolumn{1}{c|}{LBYL-Net~\cite{huang2021look}}        & \multicolumn{1}{c|}{DarkNet-53}                & 79.67          & 82.91          & \multicolumn{1}{c|}{74.15}          & {\ul 68.64}    & {\ul 73.38}    & \multicolumn{1}{c|}{{\ul 59.49}}    & 62.70          & -              & -              \\
\multicolumn{1}{c|}{TransVG~\cite{deng2021transvg}}       & \multicolumn{1}{c|}{ResNet-101}                & {81.02}    & 82.72          & \multicolumn{1}{c|}{{78.35}}    & 64.82          & 70.70          & \multicolumn{1}{c|}{56.94}          & {67.02}    & 68.67          & 67.73          \\
\multicolumn{1}{c|}{TransVG\ (Swin)}                              & \multicolumn{1}{c|}{Swin-S}                    & {\ul 82.33}          & {\ul 84.01}          & \multicolumn{1}{c|}{{\ul 79.83}}          & 64.94          & 70.19          &  \multicolumn{1}{c|}{56.47}         & {\ul 67.81}          & 69.34          & 68.99 \\  \midrule
\multicolumn{1}{c|}{\ModelName\ (ours)}                  & \multicolumn{1}{c|}{Swin-S}                    & \textbf{84.01} & \textbf{85.85} & \multicolumn{1}{c|}{\textbf{82.34}} & \textbf{72.94} & \textbf{76.17} & \multicolumn{1}{c|}{\textbf{63.81}} & \textbf{71.89} & \textbf{73.03} & \textbf{72.52} \\ 
\bottomrule

\end{tabular}
\caption{The performance comparisons (Acc@0.5) on ReferCOCO, ReferCOCO+, and ReferCOCOg. The results of previous best two-stage and one-stage methods are highlighted with underlines. We highlight our results in bold. The results demonstrate that our method outperforms all state-of-the-art one-stage and two-stage methods.}
\label{tab:main}
\end{table*}

We also present the accuracy comparison with state-of-the-art methods on ReferCOCO, ReferCOCO+, and ReferCOCOg in \Cref{tab:main}.
In ReferCOCO and ReferCOCO+ datasets, referred objects are people in ``testA" and could be common objects in ``testB".
The expressions in ReferCOCOg are much longer than the ones in other datasets.
Note that TransVG uses a ResNet-101 and a DETR encoder split from a pre-trained DETR framework. Its backbone is more powerful than a single ResNet-101.
%
%
%
%
%

%
We observe that our \ModelName\ greatly outperforms all two-stage and one-stage state-of-the-art methods.
On RefCOCO and RefCOCO+ datasets, our method gains 1.84\%$\sim$2.52\% absolute improvement in ``testA" and 2.51\%$\sim$4.32\% in ``testB".
When the referred objects are arbitrary, the inconsistent issue will be more serious, and our modulated backbone can better filter irrelevant objects and correct the representation with textual guidance.
In the RefCOCOg test split, we notice that ISRL~\cite{sun2021iterative} performs better than TransVG~\cite{deng2021transvg}.
It models the visual grounding as a Markov decision process that handles long expressions iterative by filtering out irrelevant areas.
However, the limited action space makes it easy to converge to a local optimum.
Our model does not modify the Visual-Linguistic Transformer in the TransVG and only provides query-consistent visual features.
The performance is greatly improved, which demonstrates that our query-aware refinement is more effective to model the representation for visual grounding.

\subsection{Ablation Study}

We conduct ablation studies on ReferItGame and Flickr30k to reveal the effectiveness of the proposed \ModelName.
%

%
%
We first study the effectiveness of \AttentionName\ in query-refined feature extraction and query-aware multiscale fusion.
As shown in \Cref{tab:abl_module},
we use a checkmark to denote enabling \AttentionName\ in the corresponding module, and a cross mark to denote disabling \AttentionName\ and passing the feature without modification.
When disabling \AttentionName\ in multiscale fusion, the performance is dropped by
2.52\% in ReferItGame-\textit{test} and 0.79\% in Flickr30k-\textit{test}, respectively.
When disabling \AttentionName\ in query-refined feature extraction, the performance is dropped by
3.22\% and 1.51\%.
When completely disabling \AttentionName, the performance is dropped by
3.75\% and 3.77\%.
We find that \AttentionName\ is more important in query-refined feature extraction than in multiscale fusion.
We also notice that, when compared with completely disabling \AttentionName, only enabling \AttentionName\ in multiscale fusion gains little improvement because the features from the transformer are still noise and query-inconsistent.
%
%
%
%
%

%
We further study the effectiveness of spatial attention and channel attention in the \AttentionName.
%
%
Spatial attention can filter out irrelevant areas.
And channel attention can redistribute the importance of the features to fit different query sentences.
The results shown in the \Cref{tab:abl_attention} demonstrate that spatial and channel attention are both effective.
%

\begin{table}[tb]
\small
\centering
\setlength{\tabcolsep}{4pt}
\begin{tabular}{c|cc|cc|cc}
\toprule
\multirow{2}{*}{Models} & \multirow{2}{*}{\tabincell{c}{{\scriptsize Feature} \\ {\scriptsize Extraction}}} & \multirow{2}{*}{\tabincell{c}{{\scriptsize Multiscale} \\ {\scriptsize Fusion}}} & \multicolumn{2}{c}{ReferItGame} & \multicolumn{2}{c}{Flickr30k} \\
 &  &  & val & test & val & test \\
 \midrule
\ModelName & \CheckmarkBold & \CheckmarkBold  & \bf 76.84 & \bf 74.61 & \bf 80.83 & \bf 81.95 \\
\midrule
(a) & \CheckmarkBold & \XSolidBrush & 74.31 & 72.09 & 80.09 & 81.16 \\
(b) & \XSolidBrush & \CheckmarkBold & 73.63 & 71.39 & 79.35 & 80.44\\
(c) & \XSolidBrush & \XSolidBrush   & 73.25  & 70.86 & 77.17 & 78.18\\ 
 \bottomrule
\end{tabular}
\caption{Ablation studies of \AttentionName\ in two phases of \ModelName.} \label{tab:abl_module}
\vspace{1.2em}
\small
\setlength{\tabcolsep}{4pt}
\begin{tabular}{c|cc|cc|cc}
\toprule
\multirow{2}{*}{Models} & \multirow{2}{*}{\tabincell{c}{{\scriptsize Channel} \\ {\scriptsize Attention}}} & \multirow{2}{*}{\tabincell{c}{{\scriptsize Spatial}\\{\scriptsize Attention}}} & \multicolumn{2}{c}{ReferItGame} & \multicolumn{2}{c}{Flickr30k} \\
 &  &  & val & test & val & test \\
 \midrule
\ModelName & \CheckmarkBold & \CheckmarkBold  & \bf 76.84 & \bf 74.61 & \bf 80.83 & \bf 81.95 \\
\midrule
(d) & \CheckmarkBold & \XSolidBrush & 74.35 & 72.02 & 80.22 & 81.35 \\
(e) & \XSolidBrush & \CheckmarkBold & 74.41 & 71.80 & 80.67 & 81.55\\
 \bottomrule
\end{tabular}
\caption{Ablation studies of different attentions in \AttentionName.}
\label{tab:abl_attention}
\end{table}



\subsection{Qualitative Results}
\begin{figure}[tb]
\centering
\includegraphics[width=1\linewidth]{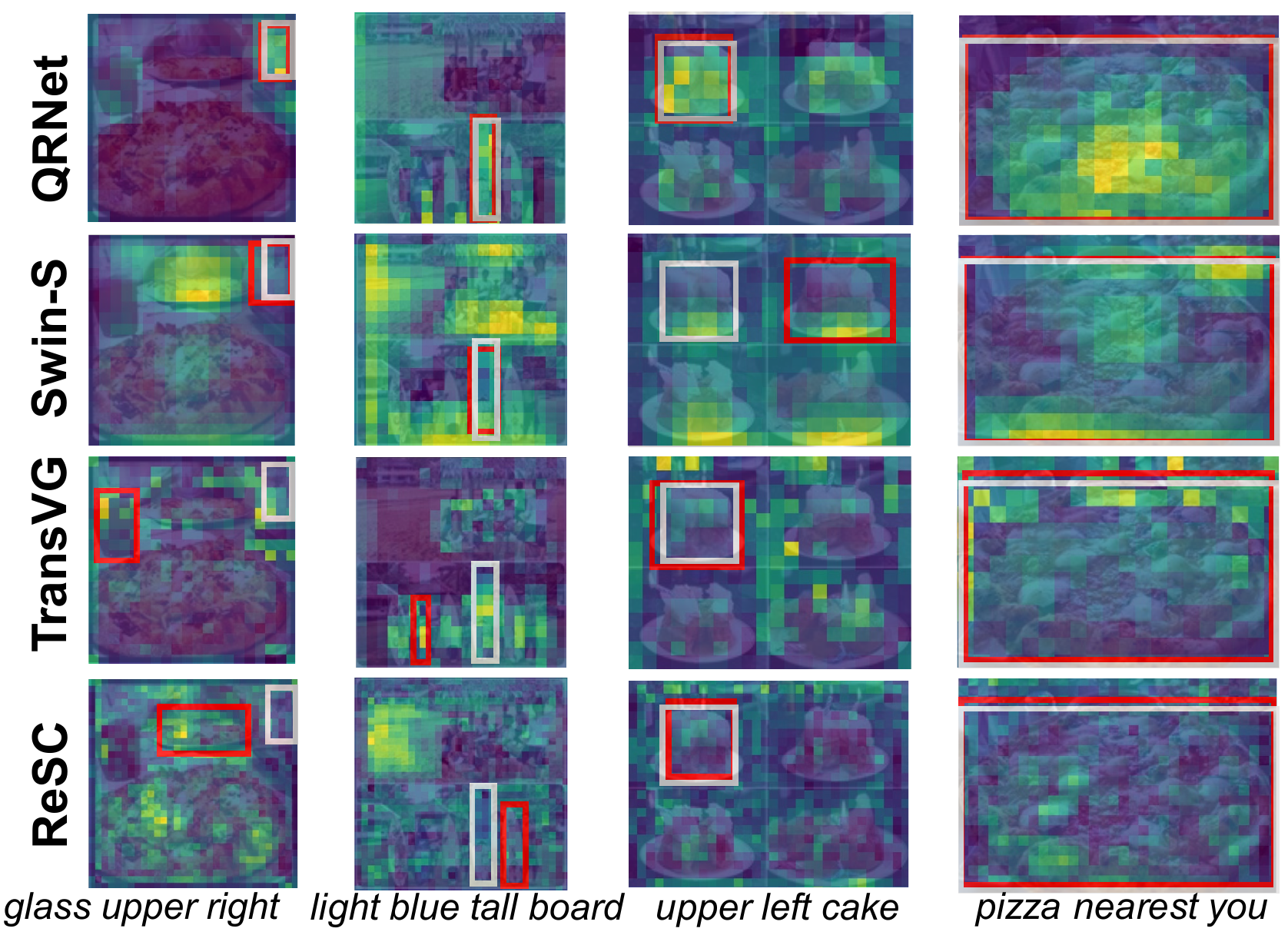}
\caption{Visualization of activation maps from the backbone of our \ModelName\ and other popular models. Red: the predicted box. White: the ground-truth box.}
\label{fig:vis}
\vspace{-1em}
\end{figure}

%
We show the qualitative comparison between our \ModelName\ and three popular methods in \Cref{fig:vis}. The feature maps are extract from the backbone of each model.
We can see that previous methods
are sensitive to many query-unrelated areas, which may lead to wrong predictions, e.g., the results of TransVG and ReSC for the queries of ``glass upper righ" and ``light blue tall board", and the result of Swin-S for the query of ``upper left cake".
In contrast, our QRNet generates query-consistent features and makes more accurate predictions. More results can be found in the supplementary.
\section{Online Deployment}
Previous experimental results have proven the advantages of our proposed QRNet, so we deploy it in a real-world online environment to test its practical performance. Specifically, we apply QRNet to enhance the search engine in Pailitao at Alibaba and perform A/B testing to evaluate the impact of our model. Details can be found in supplementary materials. We observe that QRNet decreases the no click rate by 1.47\% and improves the number of transactions by 2.20\% over the baseline. More specifically, the decrease of no click rate implies that QRNet can generate more accurate target boxes so that users are more likely to click. The improvement of the number of transactions means that the clicked item is exactly the users want to purchase, which also reveals the great performance of QRNet.

\section{Conclusion}
In this paper, we argue that pre-trained 
visual backbones cannot produce visual features that are consistent with the requirement of visual grounding.
%
%
To overcome the weaknesses, 
%
we  propose  a \NetworkName\ (\ModelName)\ to adjust  the  visual feature maps with the guidance of query text.
%
%
The \ModelName\ is designed based on a novel Query-aware Dynamic   Attention mechanism, which can dynamically compute query-dependent spatial and channel attentions for refining visual features.
%
%
Extensive experiments indicate that the improved framework significantly outperforms the state-of-the-art methods.
The proposed \ModelName\ exhibits vast potential to improve multimodal reasoning.
In future work, we plan to improve the fine-grained interaction ability of \ModelName\ and discard the post-interaction modules to simplify the existing end-to-end visual grounding frameworks.

\paragraph{Acknowledgement} We thank Zenghui Sun, Quan Zheng and Weilin Huang at Alibaba Group for their help on online deployment and the anonymous reviewers for their valuable suggestions. This work was supported by the National Innovation 2030 Major S\&T Project of China (No.2020AAA0104200 \& 2020AAA0104205), the Science and Technology Commission of Shanghai Municipality (No. 19511120200 \& 21511100100 \& 18DZ2270800), the Open Fund of PDL, and Alibaba Group through Alibaba Research Intern Program.

{\small
\bibliographystyle{ieee_fullname}
\bibliography{egbib}
}

\clearpage
\appendix

\section{Dataset Analysis}

We present the statistics about five datasets we used in our experiments.
Table~\ref{tab:dataset_st} shows the statistics.

\paragraph{ReferItGame:}
ReferItGame~\cite{kazemzadeh2014referitgame} contains images from SAIAPR12~\cite{escalante2010segmented} and collects expressions by a two-player game.
In this game, the first player is shown an image with an object annotation and asked to write a natural language expression referring to the object.
The second player is shown the same image and the written expression and asked to click the object's corresponding area.
If the clicking is correct, the two players receive points and swap roles.
If not, the new image would be presented.
\paragraph{RefCOCO, RefCOCO+, RefCOCOg:}
These three datasets contain images from MSCOCO~\cite{lin2014microsoft}.
Expressions in RefCOCO~\cite{yu2016modeling} and RefCOCO+~\cite{yu2016modeling} are also collected by the two-player game proposed in ReferitGame~\cite{kazemzadeh2014referitgame}.
There are two test splits called ``testA'' and ``testB''.
Images in ``testA'' only contain multiple people annotation.
In contrast, images in ``testB'' contain all other objects.
Expressions in RefCOCOg~\cite{mao2016generation} are collected on Amazon Mechanical Turk in a non-interactive setting.
Thus, the expressions in RefCOCOg are longer and more complex.
RefCOCOg has ``google'' and ``umd'' splits.
The ``google'' split does not have a public test set.
And there is an overlap between the training and validation image set.
The ``umd'' split does not have this overlap.
We denote these splits by ``val-g'', ``val-u'' and ``test-u'', respectively.
\paragraph{Flickr30k Entities:}
Flickr30k Entities~\cite{plummer2015flickr30k} contains images in Flickr30k dataset.
The query sentences are short noun phrases in the captions of the image.
The queries are easier to understand.
%

\section{QD-ATT in Other Vision Backbones}

The experimental results in our paper have shown that the Query-aware Dynamic Attention can help improve the transformer-based visual backbone.
In this section, we present the performance improvement in visual grounding by applying the Query-aware Dynamic Attention to DarkNet53~\cite{redmon2018yolov3} and ResNet~\cite{he2016deep} to show its generalizability in different backbones.
%
%
\paragraph{DarkNet53.}
We use a state-of-the-art one-stage visual grounding framework ReSC~\cite{yang2020improving},
which adopts the visual feature map from the 102-th convolutional layer of DarkNet53.
We append Query-aware Dynamic Attention modules after the 78-th, 91-th and 102-th convolutional layer of DarkNet53.
We take the visual feature map from the last Query-aware Dynamic Attention module as the output of the visual backbone.
\paragraph{ResNet.}
We consider the TransVG~\cite{deng2021transvg} framework, which uses ResNet50 as the visual backbone.
We apply Query-aware Dynamic Attention modules to the C2, C3, C4, and C5 layers.
The evaluation results using different visual backbones on the ReferItGame dataset are shown in \Cref{tab:qd_imp}.
The proposed QD-ATT can also significantly improve the performance.
%
%

\begin{table}[t]
  \centering
  \begin{tabular}{@{}llll@{}}
  \toprule
  Dataset            & Images & Instances & Queries  \\ \midrule
  RefCOCO            & 19,994 & 50,000    & 142,210 \\
  RefCOCO+           & 19,992 & 49,856    & 141,564 \\
  RefCOCOg           & 25,799 & 49,822    & 95,010  \\
  ReferItGame        & 20,000 & 19,987    & 120,072 \\
  Flickr30K Entities & 31,783 & 427,000   & 427,000 \\
  \bottomrule
  \end{tabular}
  \caption{Data statistics of RefCOCO, RefCOCO+, RefCOCOg, ReferItGame and  Flickr30K Entities.}
  \label{tab:dataset_st}
  \end{table}

\begin{table}[t]
  \centering
  \begin{tabular}{@{}lcc@{}}
  \toprule
  \multicolumn{1}{l|}{\multirow{2}{*}{Model}} & \multicolumn{2}{c}{ReferItGame} \\
  \multicolumn{1}{l|}{}                       & val            & test           \\ \midrule
  \multicolumn{3}{l}{\textbf{\emph{DarkNet53}}}                                                 \\ \midrule
  \multicolumn{1}{l|}{ReSC-Base}              & 66.78          & 64.33          \\
  \multicolumn{1}{l|}{\quad +QD-ATT}     & $\text{67.85}_{\text{+1.07}}$          & $\text{65.21}_{\text{+0.88}}$          \\ \midrule 
  \multicolumn{3}{l}{\textbf{\emph{ResNet50}}}                                                  \\ \midrule
  \multicolumn{1}{l|}{TransVG}                & 71.60           & 69.76          \\
  \multicolumn{1}{l|}{\quad +QD-ATT}       & $\text{75.01}_{\text{+3.41}}$          & $\text{72.67}_{\text{+2.91}}$          \\ \bottomrule
  \end{tabular}
  \caption{Performance of ReSC and TransVG when QD-ATT is adopted.}
  \label{tab:qd_imp}
  \end{table}

\section{Ablation study on RefCOCO}
We also present the ablation study results on RefCOCO dataset in \Cref{tab:sub_abl_module} and \Cref{tab:sub_abl_attention}. The experimental results reveal similar conclusions. We also notice that spatial attention is more useful in ``testA'' split and channel attention is more useful in ``testB'' split. It is because the referents in ``testB'' can be arbitrary objects but images in ``testA'' only contain multiple people annotation.

\section{Analysis of hyper-parameter K}

\begin{figure}[tb]
\centering
\includegraphics[width=1\linewidth]{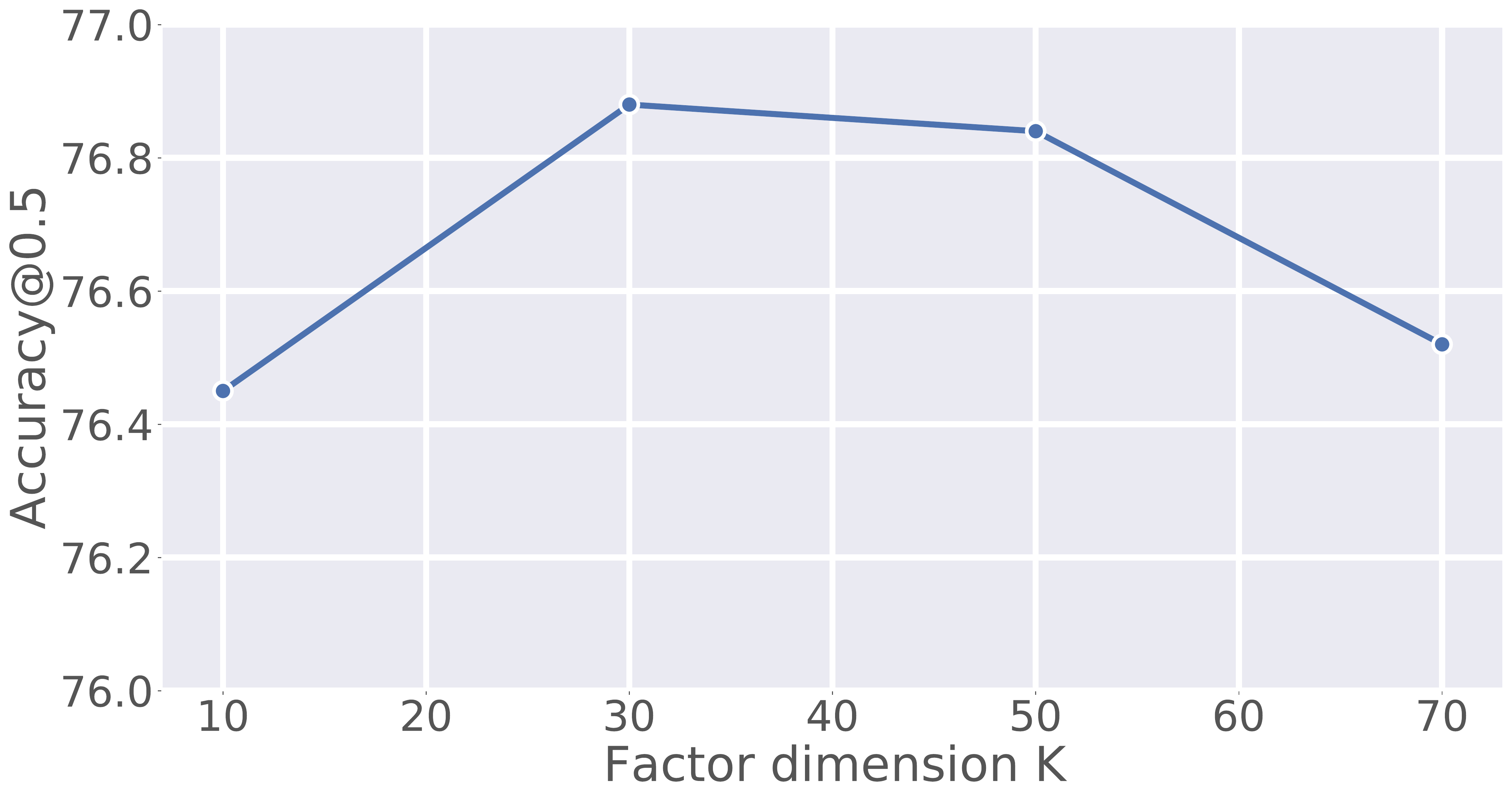}
\caption{The effect of factor dimension $K$ on the validation split of ReferItGame dataset.}
\label{fig:factor}
\vspace{-1.4em}
\end{figure}

We study the impact of hyper-parameter $K$ which denotes the factor dimension.
We report the accuracy of our model with $K=\{10,30,50,70\}$ in \Cref{fig:factor}.
When $K$ is small, the \AttentionName\ cannot learn the transformation well.
And when $K$ is large, the \AttentionName\ is easy to overfit and performs worse on validation set.
We take the best $K=30$ on the validation set as our default setting.

\section{Efficiency Analysis}

\begin{figure}[tb]
\centering
\includegraphics[width=1\linewidth]{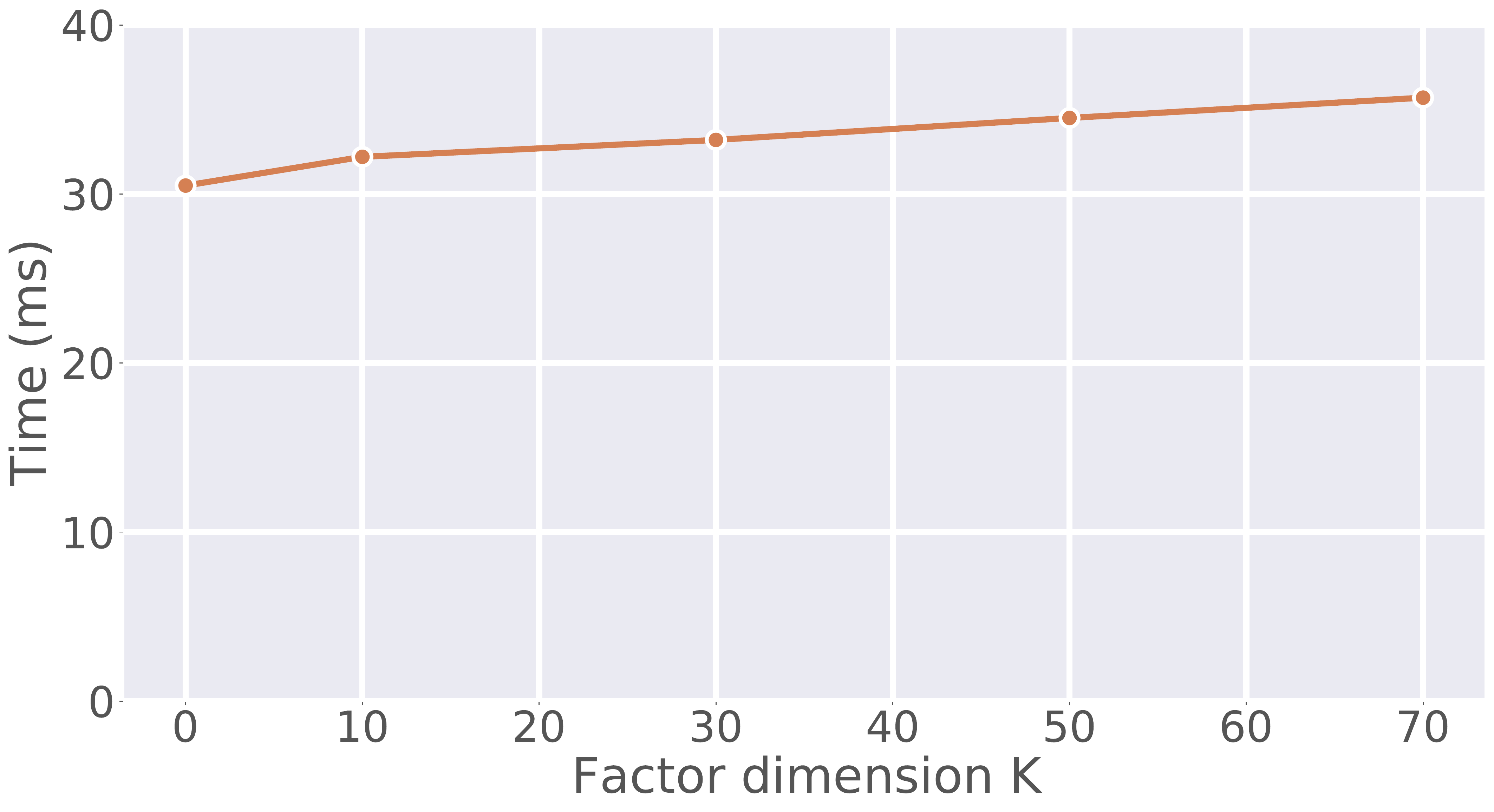}
\caption{The running time (ms) of our model with different factor dimension $K$ on the validation split of ReferItGame.}
\label{fig:factor_time}
\end{figure}

We study the running time of our model with different factor dimensions in the Query-aware Dynamic Attention.
We set $K=\{0,10,30,50,70\}$.
$K=0$ is a special setting which means the Query-aware Dynamic Attention directly returns the input feature without any computation.
We conduct the experiment on a single NVIDIA RTX3090 GPU.
The results are shown in \Cref{fig:factor_time}.
We notice that the running time increases linearly with the growth of K.
We also observe that incorporating the Query-aware Dynamic Attention does not introduce excessive computation cost, which means our QD-ATT is efficient and effective.

\begin{table}[t]
  \centering
      \begin{minipage}{1\linewidth}
        \centering
        \small
          \begin{tabular}{c|cc}
              \toprule
              &\multicolumn{2}{c}{RefCOCO}\\
               & testA & testB  \\
               \midrule
              (a) & 84.63 & 81.42  \\
              (b) & 84.46 & 80.96  \\
              (c) & 84.01 & 79.83 \\ 
               \bottomrule
          \end{tabular}
      \caption{Ablation studies of QD-ATT in two phases of QRNet.} 
      \label{tab:sub_abl_module}
      \vspace{1em}
      \end{minipage}%
      \hspace{.01\linewidth}
      \begin{minipage}{1\linewidth}
        \centering
        \small
          \begin{tabular}{c|cc}
              \toprule
              &\multicolumn{2}{c}{RefCOCO} \\
               & testA & testB \\
               \midrule
              (d) & 84.61 & 81.25 \\
              (e) & 85.43 & 80.91 \\
           \bottomrule
          \end{tabular}
      \caption{Ablation studies of different attentions in QD-ATT.}
      \label{tab:sub_abl_attention}
      \end{minipage} 
  \end{table}
  
\section{Qualitative Analysis}

\begin{figure*}[htb]
  \centering
  \includegraphics[width=\textwidth]{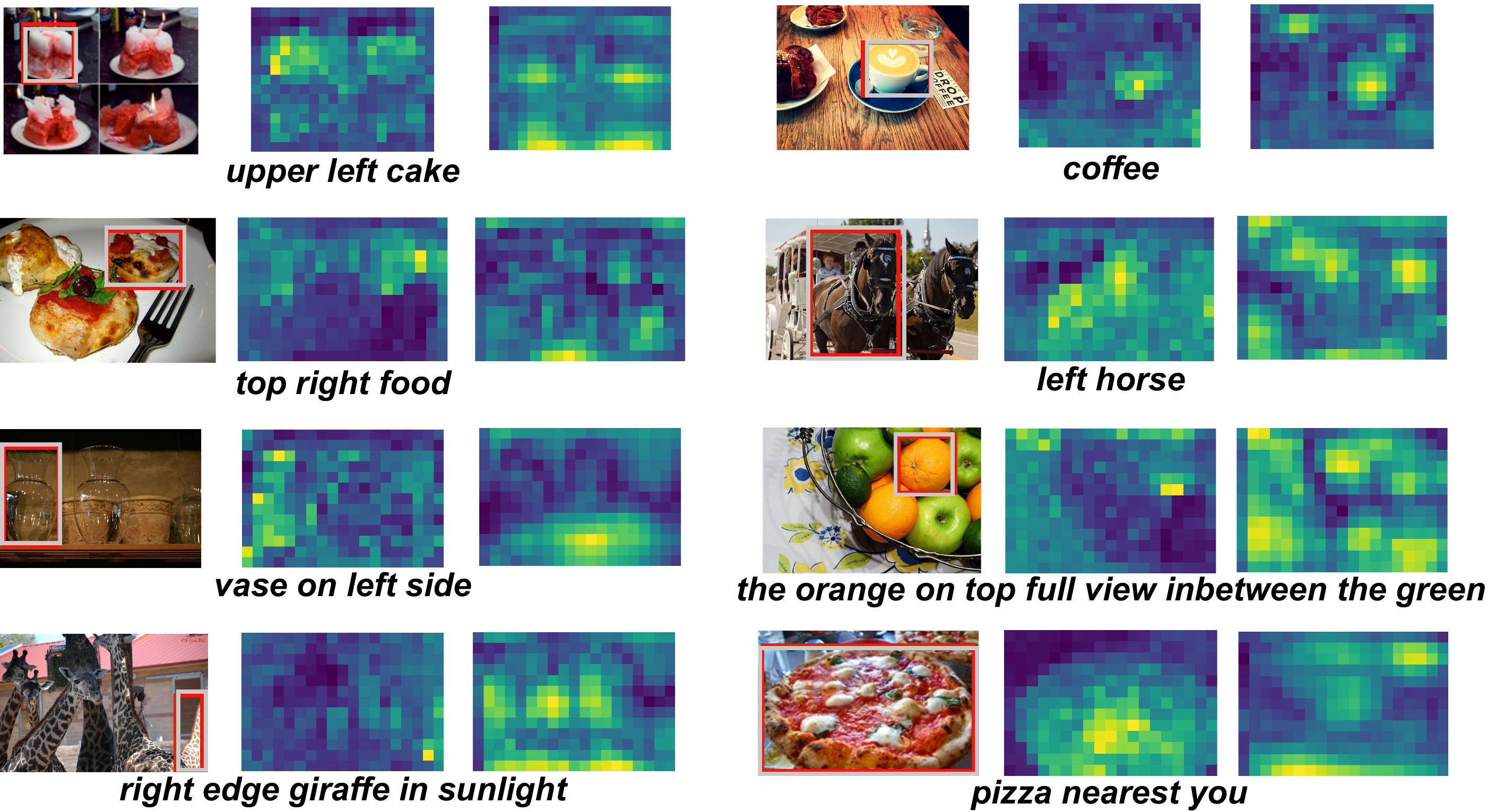}
  \caption{Visualization of activation maps from the backbone of our \ModelName\ and original Swin-Transformer. Red: the predicted box. White: the ground-truth box.}
  \label{fig:vis_wide}
  \end{figure*}

\paragraph{Feature Refinement Visualization.}
We present more visualize cases in \Cref{fig:vis_wide}. It shows the ability of our QRNet to refine general purposed feature maps to query-aware feature maps.

\begin{figure*}[b]
  \centering
  \includegraphics[width=0.8\linewidth]{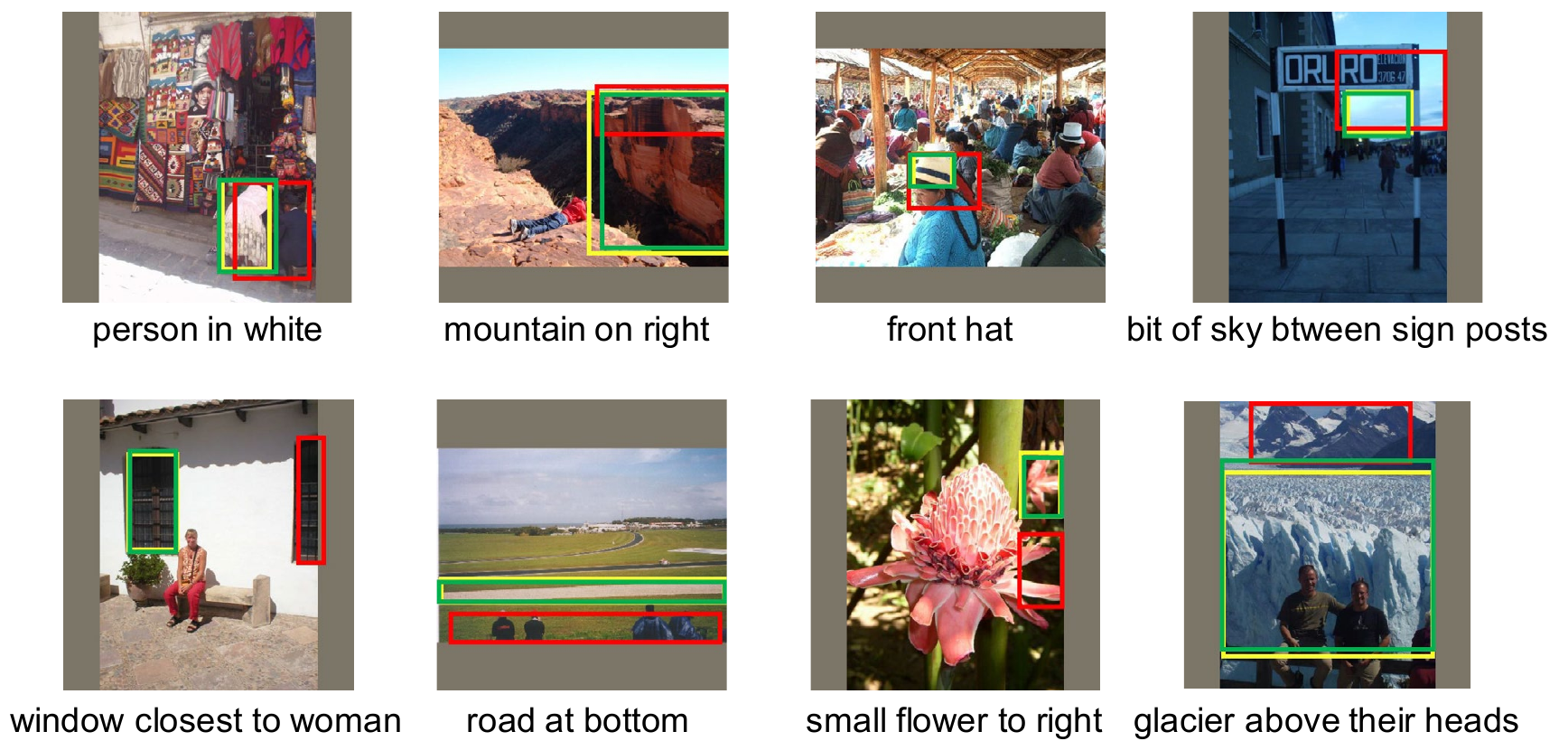}
  \caption{
  Qualitative comparison with TransVG.
  Green boxes are the ground-truth, yellow boxes are the predictions of our model, and red boxes denote the predictions of TransVG.}
  \label{fig:case}
  \end{figure*}

\paragraph{Comparison with TransVG.}
We quantitatively compare the results predicted by our model and the TransVG in \Cref{fig:case}.
Green boxes are the ground truth.
Yellow and red boxes are the predictions of our model and TransVG, respectively.
In the first row, we notice that the predictions of TransVG are close to the ground truth.
However, sometimes the TransVG only localizes the significant areas of the target objects (e.g., ``mountain on right"), and sometimes the predicted boxes cover the irrelevant areas around the target objects (e.g.,  ``person in white"), resulting in failures.
In the second row, the TransVG does not fully capture the semantics of queries and is vulnerable to the interference of significantly noisy objects in the image, resulting in incorrect predictions.
Our model utilizes query sentence representation to guide the visual backbone to obtain query-aware visual features, which helps the framework give correct predictions.

\begin{figure*}[b]
\centering
\includegraphics[width=0.8\linewidth]{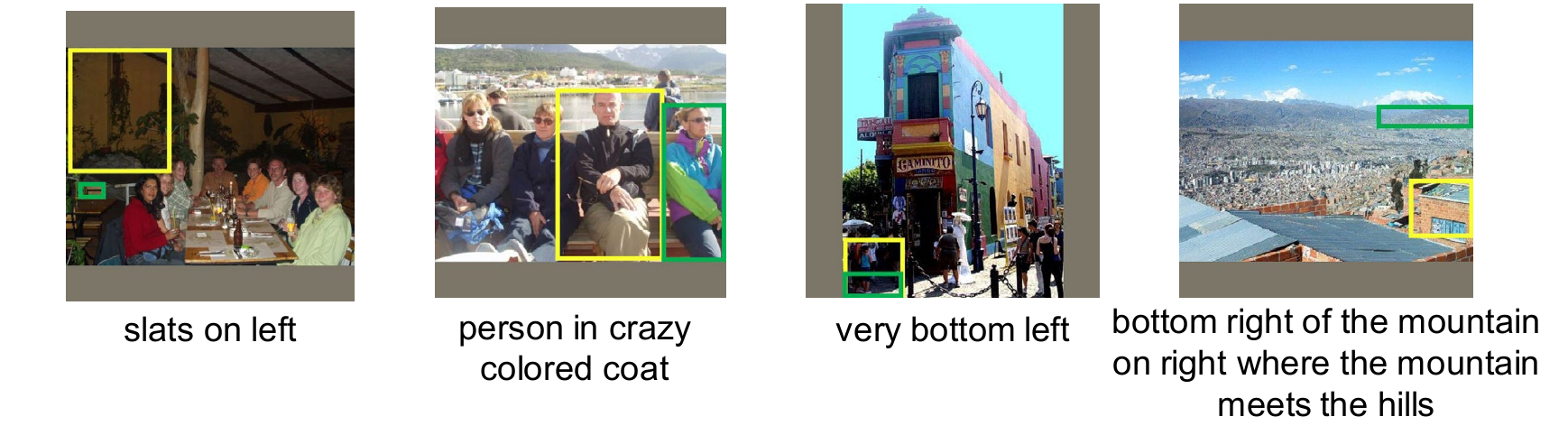}
\caption{The failure cases of our model. Green boxes are the ground-truth, yellow boxes are the predictions of our model.}
\label{fig:failure_case}
\end{figure*}

\paragraph{Failure Cases.}
We show some typical failure cases in \Cref{fig:failure_case}.
One type of error is the long tail description.
For the first example, our model does not recognize ``slats'' correctly.
During both the pre-training the fine-tuning, the samples of long-tail concepts are infrequent.
Thus, our method fails.
For the second image, we notice that ``crazy'' is an abstract word, and its visual appearance can be completely different in different contexts.
Therefore, it is also hard for our model to learn.
In the third example, we found that our model understands the query's intention, but it tends to predict a box containing some objects, and the annotator does not have such bias.
For the last example, 
we find that our model may make mistakes when facing long sentence queries.
Our model only takes the [CLS] representation rather than the complete sequence representation from BERT to refine the visual features, limiting the ability to understand long text.
%

\section{Limitation}
In this paper, we propose \AttentionName\ to extract query-consistent features from the visual backbone.
However, we only take the contextual textual representation of the query to keep it efficient.
Due to the lack of fine-grained multimodal interaction, we still need a post-interaction module in the visual grounding framework.
It also hinders generalizing our module to the task requiring complex reasoning, \eg, multimodal dialogue~\cite{nie2019multimodal} and visual storytelling~\cite{huang2016visual}.

\section{Online Deployment Setting}
We apply QRNet to enhance the search engine in Pailitao at Alibaba and perform A/B testing. In order to reduce the discrepancy between the images from consumers and sellers, it is important to locate the target from the item image. Note that a  detection-based two-stage method is already deployed online.
The A/B testing system split online users equally into two groups and direct them into two separate buckets, respectively. Then in bucket A (i.e., the baseline group), the target boxes from images are generated by a detection-based method. While for users in the B bucket (i.e., the experimental group), the target boxes are generated by the proposed QRNet.
All the conditions of the two buckets are identical except for the target grounding methods.

We calculate the clicked query rate and the number of transactions for all the categories in the two buckets. 
The online A/B testing lasts for two days, and over 1.5 million daily active users are in each bucket. 
We find that the performance in the experimental bucket (i.e., QRNet) is significantly better (p $<$ 0.05) than that in the baseline bucket (i.e., detection-based) on both measures. QRNet decreases the no click rate by 1.47\% and improves the number of transactions by 2.20\% over the baseline. More specifically, the decrease of the no click rate implies that QRNet can generate more accurate target boxes so that users are more likely to click. The improvement of the number of transactions means that the clicked item is exactly what the users want to purchase, which also reveals the great performance of QRNet.

\section{Pseudo-code of QD-ATT}
We present a Pytorch-style pseudo-code in Listing~\ref{lst:qda}.
\begin{figure*}
\begin{lstlisting}[language=Python, caption=Pytorch-style pseudo-code of Query-aware Dynamic Attention, label={lst:qda},xleftmargin=.05\textwidth, xrightmargin=.05\textwidth]
# Perform linear transform to change the dimension of the vector space from in_dim to out_dim
# Inputs:
#   F: the input visual features (B, H, W, D_in)
#   q: the input linguistic feature (B, D_q)
# Return:
#   F_q: the output transformed visual features (B, H, W, D_out)
# Parameters:
#   linear: a plain linear layer to generate factor U
#   S: a random-initialized factor matrix (K, D_out)
def dynamic_linear(F, q):

  # Parameter generation
  # K is the  factor dimension

  # generate the factor matrix U
  U = reshape(linear(q), (B, D_in+1, K))
  # reconstructed matrix by U x S
  M = einsum("bik,kj->bij", U, S)  
  Weight, bias = M[:,:-1], M[:,-1]

  # Perform linear transform
  F_q = einsum("b...d,bde->b...e", F, Weight) + bias
  return F_q

# Perform dynamic attention on visual feature map by textual guidance.
# Inputs:
#   F: input visual feature map (B, H, W, C)
#   q: [CLS] token's last hiddenn states from BERT (B, D_q)
# Return:
#   F_refined: refined visual feature (B, H, W, C)
def query_aware_dynamic_attention(F, q):
    # Channel Attention
    B,H,W,C = F.shape
    F_flatten = reshape(F, (B, -1, C))
    avg_pool = mean(F_flatten, dim=1)
    # The mlp transform the channel dimension  C -> C/16 -> C
    att_avg = dynamic_linear(relu((dynamic_linear(avg_pool, q))), q)
    max_pool = max(F_flatten, dim=1)
    att_max = dynamic_linear(relu((dynamic_linear(max_pool, q))), q)
    attn = sigmoid(att_avg + att_max)
    F = F * reshape(attn, (B, 1, 1, C))
    # Spatial Attention
    # dynamic_linear transform channel dimension C -> 1 
    attn = sigmoid(dynamic_linear(F, q))
    F_refined = F * attn
    return F_refined
\end{lstlisting}
\end{figure*}




\end{document}